# Clustering Guided Domain-Specific Pretrained Foundation Model for Very High-Resolution Arctic Remote Sensing

Amal S. Perera[1], *Member IEEE*, Chandi Witharana[1], Elias Manos[1], Michael Pimenta[1], Anna K. Liljedahl[3].

***Abstract*–This study introduces a novel Arctic-focused remote sensing foundation model (RSFM) by combining diversity-aware regional-scale image curation with masked autoencoder (MAE) self-supervised pretraining of a vision transformer (ViT) encoder for very high spatial resolution (VHSR) satellite image analysis tasks. RSFMs aim to learn transferable visual representations from large, diverse Earth observation archives using self-supervision, enabling greater adaptation to downstream tasks with limited labeled data. In this study, spectral and acquisition-metadata descriptors were used within a scalable affinity-propagation clustering workflow to select approximately 3 million representative training patches from 267 TB of Vantor VHSR multispectral satellite imagery spanning over 5 million km² of Arctic permafrost tundra ecoregion. This curation strategy was designed to reduce oversampling of visually repetitive or low-information areas while preserving broad scene diversity across the study domain. We pretrained a ViT-Large encoder on the curated corpus using a domain-adapted MAE reconstruction objective designed for VHSR satellite images. This pretraining produced Arctic-specific transformer weights for downstream feature mapping tasks. The pretrained encoder was integrated into an existing, detection and segmentation framework augmented with location embeddings, and evaluated across four hand-labeled Arctic feature-mapping datasets: human-built infrastructure, ice-wedge polygons (IWP), retrogressive thaw slumps (RTS), and tundra capillary networks (TCNs). Compared with an ImageNet-initialized ViT-Large baseline, Arctic MAE pretraining produced consistent improvements in foreground mean F1 scores of 0.87, 0.72, 0.93, and 0.87, respectively for infrastructure, IWP, RTS, and TCNs, with approximately 5–8 percentage increase across the evaluated tasks. The proposed model also outperformed Prithvi-EO-2.0, Earth foundation model in all downstream comparisons achieving at least 15% percentage improvement, suggesting that domain-specific self-supervised pretraining on curated Arctic VHSR imagery provides more transferable representations for fine-scale Arctic mapping than a general-purpose Earth observation foundation model. Our results demonstrate that optimizing the pretraining data distribution at regional scale, while keeping the architecture and MAE objective fixed, can produce a reusable Arctic-domain encoder for multiple VHSR remote sensing applications.**

This research was supported by the Google.org's Impact Challenge on Climate Innovation grant, U.S. National Science Foundation (NSF) grant #: 1720875, 1722572, 1927872, 1927723, 1927729. eXtreme Science and Engineering Discovery Environment (Award #: DPP 190001) and Texas Advanced Computing Center Award #: DPP20001). Authors would like to thank Polar Geospatial Center, University of Minnesota for imagery support under NSF-OPP awards 1043681 and 1559691 (Corr. author Amal S. Perera).

[1] Department of Natural Resources and the Environment, University of Connecticut, Storrs, CT. amal.perera, chandi.witharana, elias.manos, michael.pimenta@uconn.edu.

[3] Woodwell Climate Research Center, Falmouth, MA., aliljedahl@woodwellclimate.org.

## I. INTRODUCTION

Environmental systems are undergoing rapid and non-linear change driven by climate warming, land use pressures, and coupled biophysical feedback [1], [2], [3]. From tropics to high latitude Arctic boreal and tundra systems, landscapes now exhibit abrupt transitions and recovery dynamics that demand for innovative monitoring approaches from local to global scales [4], [5]. Over the past several decades, we have entered an era of unprecedented Earth observation (EO) data availability from ground-based sensing, conventional aerial platforms, low altitude drone systems to space-borne platforms. Historical aerial archives dating back to early 1930s, multi-decadal moderate resolution satellite missions, such as MODIS, Landsat, and Sentinel , and newer sub-meter resolution commercial satellite constellations (e.g., Vantor, Planet) collectively provide continuous, multi-scale observations of the Earth system [6], [7], [8]. Emerging sensor fusion products further expand EO data ecosystem by increasing temporal frequency, spatial resolution, and spectral richness [9]. EO advances offer unprecedented potential to address long standing scientific questions related to landscape dynamics, ecosystem processes, and environmental change. However, despite the scale and diversity of available data, much of EO data archives remain underutilized, and many fundamental earth science questions remain unanswered [10]. This knowledge gap highlights a central challenge of earth informatics, which is how to transform large, heterogeneous remote sensing datasets into science-ready products that inform downstream synthesis efforts.

Recent advances in artificial intelligence, especially large scale self supervised learning and foundation models, have begun to reshape how complex data are represented and interpreted [11]. Analytical approaches such as masked autoencoders (MAEs) enable models to learn generalized representations directly from raw, unlabeled data by reconstructing missing information [12] [13]. These methods have demonstrated strong performance in computer vision (CV) and suggest a pathway toward scalable and transferable learning in EO data analysis [14]. However, the direct translation of such approaches into remote sensing contexts is not yet straightforward.

Remote sensing imagery differs fundamentally from everyday images in both structure and semantics. Geo-objects, for instance wetland, floodplain, and disturbance are realized as spatially and temporally contextualized Earth-surface patterns (fields) whose meaning arises from scale, location, and environmental relationships rather than from discrete visual form [15] [16] [17]. Everyday objects, such as a chair, car, dog, on the other hand are discrete entities recognizable by shape and texture. Geo-objects differ from everyday objects because their semantics are emergent properties of scale, spatial context, and temporal process rather than bounded visual instances[18] [19] [20]. This unique scale-constructed meaning challenges direct transfer of deep learning models (e.g. ImageNet-style) that are pioneered in CV tasks to EO data analysis[21] [22]. Further, remote sensing data are inherently multi-spectral, multi-temporal, and influenced by sensor specific characteristics, such as viewing geometry [23]. These properties challenge conventional CV assumptions and limit the effectiveness of models trained on standard image datasets. Consequently, a compelling question we often come across is that can a single generalized deep learning model adequately represent diverse Earth system processes, or is domain adaptation necessary to capture environment specific patterns and dynamics?

Remote sensing foundation models (RSFMs) can be viewed as a specialized subset of geospatial foundation models (GFMs), focusing primarily on learning representations from Earth observation imagery, whereas GFMs aim to integrate multi-modal geospatial data for broader spatial reasoning tasks. Recent RSFMs including transformer-based architectures and masked autoencoder variants, such as Prithvi [24], [25], [26], [27], SatMAE, and Scale-MAE, have begun to stimulate the momentum in the Geospatial Artificial Intelligence (GeoAI)[28]. RSFMs attempt to address automated remote sensing image analysis challenges by capitalizing on large scale EO datasets harnessed with representation learning mechanisms. While current RSFMs show promise across a range of downstream EO image data tasks via fine tuning (few-shot or zero-shot learning mechanisms), they are primarily trained on medium to coarse resolution satellite imagery or mixtures of moderate and high-resolution data. Consequently, performance of RSFMs is often constrained by limitations in training data selection and representativeness. Remote sensing data represent a distinct and inherently multi-modal domain, characterized by heterogeneous sensing mechanisms, multi-scale spatial structure, and complex physical acquisition processes; inadequate consideration of spatial texture, scale-dependent structure, and process-driven variability further constrain the ability of RSFMs to generalize across regions and applications [29], [30], [31], [32].

The Arctic could serve as an intriguing model system to experiment with the degree of generalizability of RSFMs for downstream analysis tasks. Characterized by strong environmental gradients, diverse landforms, and rapid climate-driven change, the Arctic tundra ecoregion spans over 5 million km² and exhibits pronounced heterogeneity across spatial and temporal scales [33], [34]. Embedded in a mosaic of tundra vegetation types, such as sedge, tussock, shrub, and barren. [33], permafrost landforms (e.g., atypical ice-wedge polygonal network [35], and dynamic thermokarst formations (e.g., retrogressive thaw slumps, [36] from sub-meter scale to kilometer scales present unique challenges for automated remote sensing analysis due to their subtle spectral signatures, complex morphometry, gradual and abrupt change in morphology, phenology, and strong coupling with hydrological and thermal processes [37].

Rapid Arctic warming is triggering widespread permafrost thaw-induced disturbances that impact hydrology [35], vegetation dynamics [38], [39], and biogeochemical fluxes [40], [41], while threatening the human-built environment through damage to infrastructure on permafrost-affected ground [42], [43]. Mapping and tracking land surface changes without compromising spatial details and geographical extent inherently rely on high spatial resolution satellite images. Over the years Vantor commercial satellites (e.g., Worldview 1/2/3) have repeatedly imaged entire Arctic domain ($60^0$N), accumulating >10 million image scenes. Special licensing agreements allow US NSF polar program funded researchers to freely access Vantor images via the Polar Geospatial Center. Despite the accessibility to data and cyberinfrastructure for research use, Arctic Vantor image archives remain largely unexplored, and many fundamental Arctic permafrost scientific questions remain poorly answered. RSFMs can make Vantor image archives discoverable. However, a *one-size-fits-all* path is unlikely to provide an optimal solution. We hypothesize that RSFMs trained for ecoregion-specific conditions can outperform globally trained RSFMs.

The central goal of our study is to develop a novel domain-adapted masked autoencoder framework designed specifically for Arctic remote sensing applications. In our experimental design we use tundra ecoregion as a representative testbed and capitalize on Vantor satellite imagery. Our specific objectives are three-fold, which are to; 1) develop a scalable, diversity-aware sampling strategy for curating representative Arctic tundra image patches from large satellite image archives; 2) design and train a domain-adapted masked autoencoder framework that learns Arctic-specific visual representations from very high spatial resolution imagery; and 3) evaluate whether the resulting Arctic RSFM improves downstream performance and transferability across representative permafrost-related remote sensing tasks compared with globally trained foundation models and task-specific baselines.

## II. METHODS

Our methodological framework was designed to develop an Arctic RSFM tailored to VHSR satellite imagery and to assess its utility for downstream Arctic feature detection. In this work, we combined large-scale data curation with MAE pretraining to produce domain-specialized transformer representations for Arctic landscapes. A central challenge in foundation-model development for VHSR remote sensing is constructing a pretraining corpus that is both large enough to capture landscape diversity and selective enough to avoid overwhelming the model with redundant or uninformative imagery. To address this, we extended the ISOSCELES (Iterative Self-Organizing SCene-LEvel Sampling [44] pipeline into a scalable scene-selection framework suitable for regional-scale Arctic imagery collections. The curated corpus was then used to pretrain a Vision Transformer Large (ViT-Large) encoder using MAE training recipes adapted for VHSR satellite data. Through self-supervised reconstruction of masked image patches, the model learns spatial and spectral structures characteristic of Arctic environments, yielding pretrained weights specialized for this domain. To evaluate the utility of these learned representations, the pretrained encoder was integrated into an existing downstream feature detection framework[45], a transformer-based detection and segmentation architecture augmented with location embeddings. The resulting model was assessed across four hand-labeled Arctic feature-detection datasets to measure the extent to which Arctic-specific pretraining improves downstream performance. Overall methodological framework is shown in Figure 1.

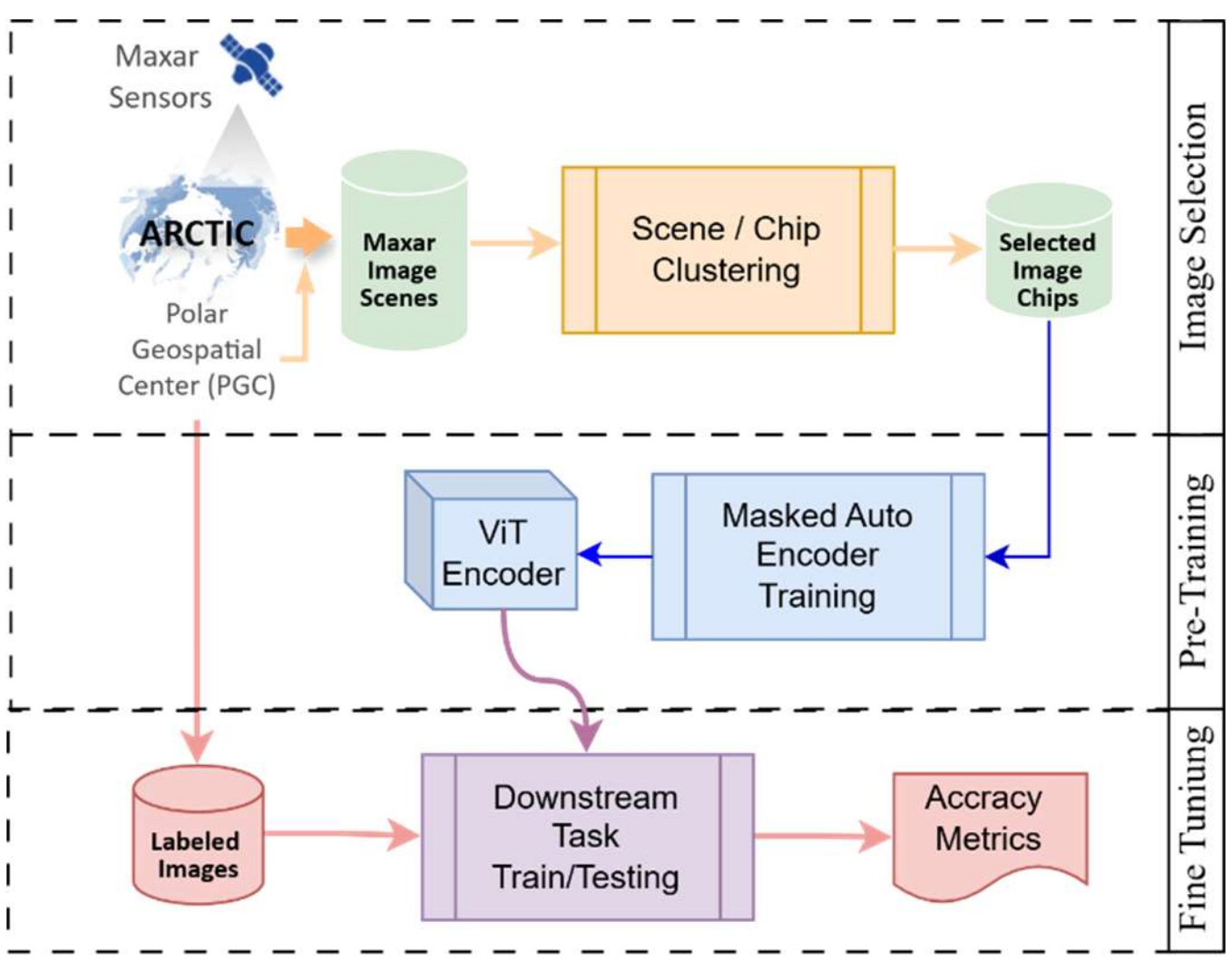

**Figure 1** Overall methodological framework, that goes through the phases of image selection, pre-training, and fine tuning for the downstream task.

The remainder of this section details each component of the workflow: (1) the Arctic image archive and downstream evaluation datasets, (2) large-scale scene characterization and representative sample selection using the extended ISOSCELES pipeline, (3) MAE-based pretraining of the ViT-Large backbone on the curated Arctic corpus, (4) integration of the pretrained encoder into the downstream detection framework, and (5) downstream evaluation across multiple Arctic detection and segmentation tasks.

### *A. Study area and Data*

The study area lies within the Arctic tundra ecoregion, spanning regions north of approximately 60°N and underlain by continuous to discontinuous permafrost [46]. The landscape is characterized by low-relief terrain, polygonal ground features, and a mosaic of wetlands, thaw lakes, and vegetated surfaces dominated by mosses, lichens, sedges, and dwarf shrubs [33] The tundra experiences long, cold winters and short, cool summers, with a brief growing season and pronounced seasonal dynamics (Figure 2). For downstream task evaluation of Arctic feature-detection, four candidate datasets were adopted from prior studies: retrogressive thaw slumps (RTS) [36], Arctic infrastructure [47], ice-wedge polygons (IWP) [45], [48], trough capillary networks (TCN) [49]. These datasets were created from sub-meter Vantor satellite imagery provided by the Polar Geospatial Center, with target features manually digitized from summer images with minimal cloud cover. The original train, validation, and test splits were retained to support comparison with prior baselines.

The RTS dataset contains manually digitized retrogressive thaw slump features from Banks Island, Northwest Territories, and the Eureka Sound Lowlands, Nunavut, Canada. The dataset includes 950 RTS objects and 2,132 image chips at 1024 × 1024 pixels. The Arctic infrastructure dataset includes buildings, roads, and storage tanks digitized from Vantor imagery across 18 sites in Alaska, Canada, and Russia. The dataset captures variability across rural, urban, industrial, tundra, and boreal settings and includes 5,374 image chips at 256 × 256 pixels. The IWP dataset contains manually digitized ice-wedge polygons from Arctic Alaska, Canada, and Russia. Polygons were labeled as low-centered or other based on morphometry. The dataset includes 33,091 labeled IWP instances across 855 image chips with variable chip sizes ranging from 199 x199 to 506 x 506. The TCN dataset contains manually digitized trough capillary networks from Jago River, Alaska. The dataset includes 2016 image chips at 1024 × 1024 pixels.

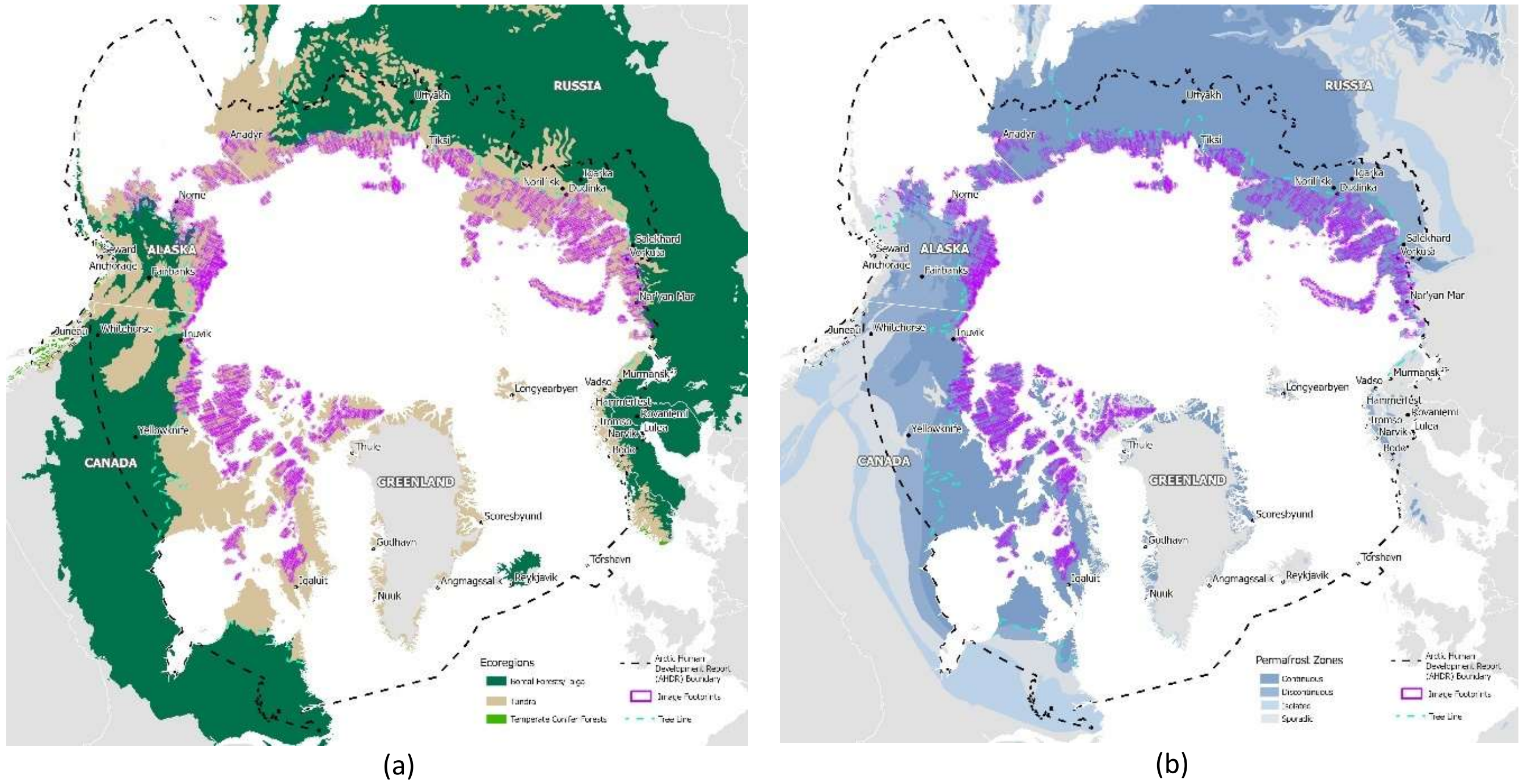

**Figure 2** Ecological and cryogenic context of the study domain. Candidate Vantor satellite imagery footprints (> 30,000) overlain on; (a) RESOLVE ecoregion map [50] and (b) permafrost zone map [46] (Brown et al. 1997).

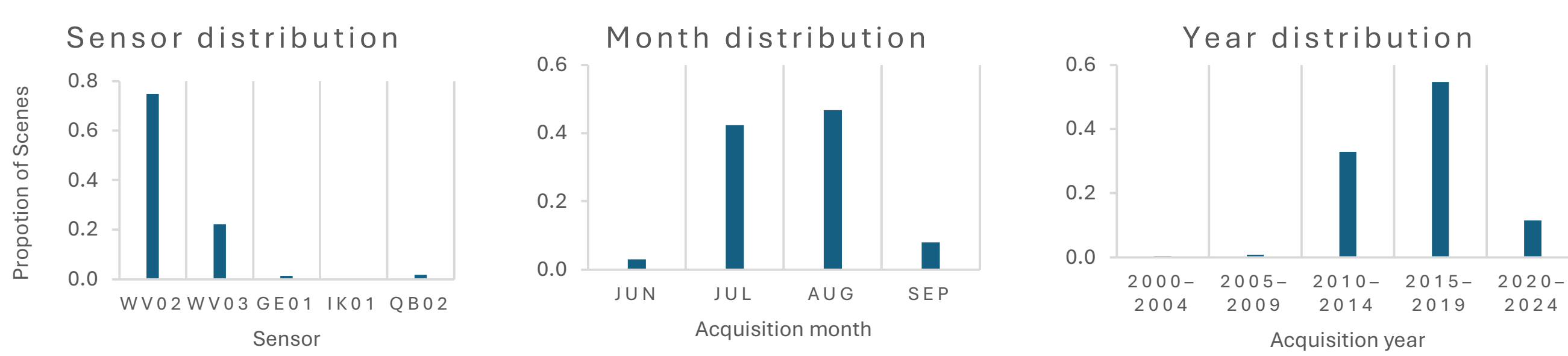

**Figure 3** Image Archive Composition (Sensor, Month, Year) of > 30,000 Vantor image scenes, showing that the corpus is dominated by the short Arctic summer acquisition window

*B. Representative Sample Selection*

To construct a representative pretraining corpus from the large Arctic imagery archive, approximately 32,000 Vantor candidate scenes (the size of each scene ~ 20 km x 20 km (40,000 x 40,000 px)) were characterized using spectral descriptors and image-level metadata and processed through the scalable ISOSCELES scene-level sampling pipeline [44] (Figure 5). Affinity-propagation clustering was used to identify representative scene exemplars without requiring the number of clusters to be specified a priori. The clustering process was controlled by the preference parameter, p, which was tuned experimentally to adjust the number of selected exemplars and, consequently, the size of the retained image-patch set. Image patches were then extracted from the selected exemplar scenes to form the final MAE pretraining dataset. Candidate scenes were represented using a reduced set of spectral and acquisition descriptors, including per-band mean and standard deviation statistics, viewing geometry, and sensor-related metadata. This feature set was obtained by screening a broader candidate descriptor pool and removing highly correlated variables prior to clustering. The reduced descriptor set was then held fixed across multiple affinity-propagation runs while tuning the preference parameter, p, to achieve the target number of exemplars. All retained variables were standardized before clustering to ensure comparability across heterogeneous attributes.

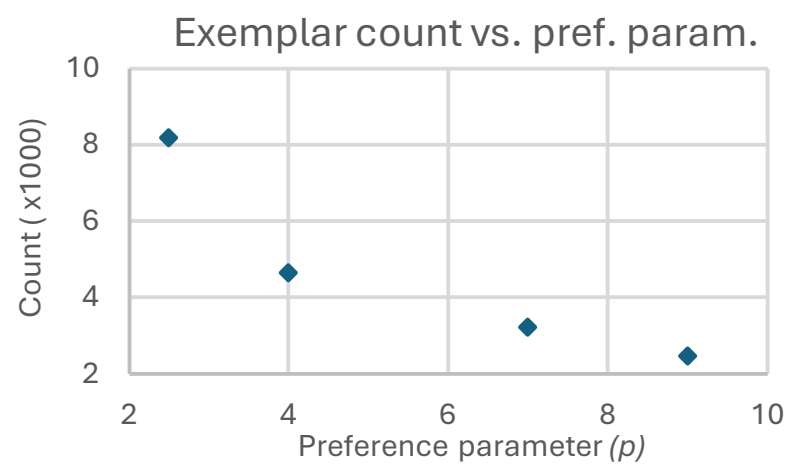

**Figure 4** Relationship between the affinity-propagation preference parameter and the number of selected exemplars. Increasing the preference parameter reduced the number of exemplars, and the final setting was selected empirically to achieve the target training-patch volume.

To observe the geographic and ecological coverage of the selected pretraining corpus, we overlaid the selected image footprints on the Circumpolar Arctic Vegetation Map (CAVM) map data on Figure 6. The CAVM provides a circumpolar reference for major tundra vegetation and land-cover types and allows the selected scenes to be interpreted relative to broad Arctic environmental gradients. The overlay demonstrates whether the selected image corpus spans multiple tundra settings, rather than being concentrated in a narrow geographic or ecological subset. This is important for MAE pretraining because representation learning from VHSR imagery depends not only on the number of image patches but also on the diversity of surface conditions, morphometry, land phenology, vegetation structure, hydrology, terrain, and sensor-acquisition contexts represented in the training data.

In the second phase, each of the previously selected 8132 exemplar scenes were partitioned into non-overlapping image chips (1024 x 1024 px), excluding chips that contained no-data pixels. Each chip was characterized using per-band mean and standard deviation statistics, together with Gabor-based texture features computed from a grayscale representation of the chip. The resulting chip descriptors were standardized by feature group and clustered using affinity propagation to identify representative chip-level exemplars within each scene. The selected exemplar chips were then exported as georeferenced GeoTIFF patches and used to construct the final MAE pretraining corpus.

This two-stage selection strategy allowed the workflow to preserve diversity at both the scene and within-scene levels. Scene-level clustering reduced redundancy across the regional archive, while chip-level clustering identified representative local surface patterns within each selected scene. This was particularly important for VHSR Arctic imagery, where a single scene may contain heterogeneous mixtures of tundra vegetation, surface water, bare ground, patterned ground, shadows, and anthropogenic features. By selecting exemplar patches within each scene, the final pretraining corpus better captured fine-scale spectral–textural variability while avoiding excessive sampling of visually redundant areas. Figure 7 shows a few example chip-level exemplar selections within selected image scenes.

### C. *Pretraining*

The representative image chips selected through the clustering-based curation pipeline was used for self-supervised pretraining using a masked autoencoder (MAE) framework [12]. A Vision Transformer Large (ViT-Large) backbone was adopted as the encoder and trained to reconstruct masked image content from the subset (0.3) of visible tokens, thereby learning transferable visual representations from unlabeled Arctic VHSR imagery. To better reflect remote-sensing spectral behavior, the reconstruction objective (Eq1) combined a covariance-aware Mahalanobis term (Eq2) with a Spectral Angle Mapper (SAM) term (Eq3), thereby capturing both multivariate spectral deviation and spectral-shape consistency [51], [52], [53], [54], [55].

Pretraining was implemented within a Detectron2-based distributed training framework and executed on a high-performance computing system using 16 compute nodes with 4 GPUs per node [56] . To identify the most effective pretrained encoder, we ran several MAE pretraining sessions in parallel using different configuration settings. During training, intermediate checkpoints were saved and selected checkpoints were evaluated through downstream fine-tuning experiments. The final pretrained model checkpoint was chosen based on its transfer performance on the downstream validation tasks, rather than on MAE reconstruction loss alone.

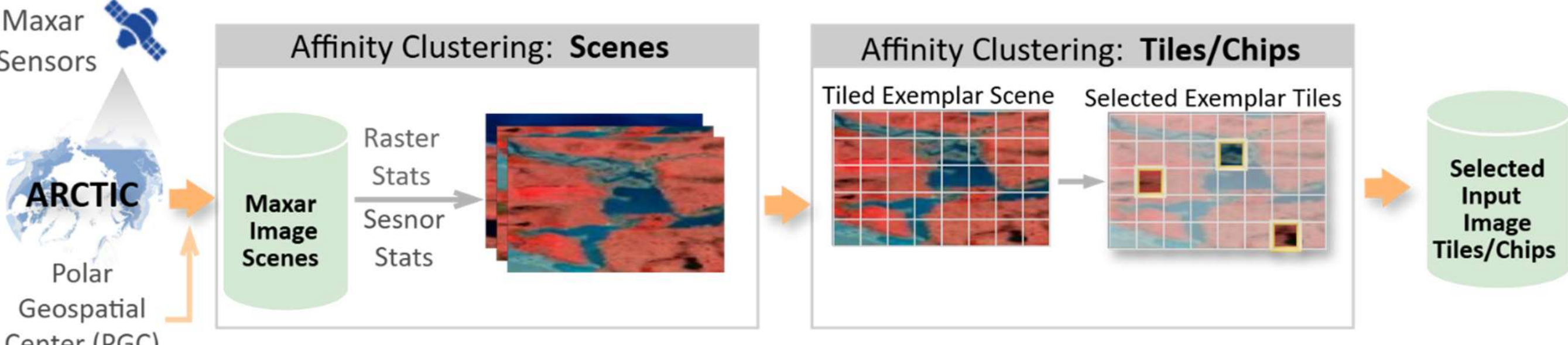

**Figure 5** Two phase ISOSCLES clustering pipeline that takes large collection of input scenes and produces a collection of representative image chips.

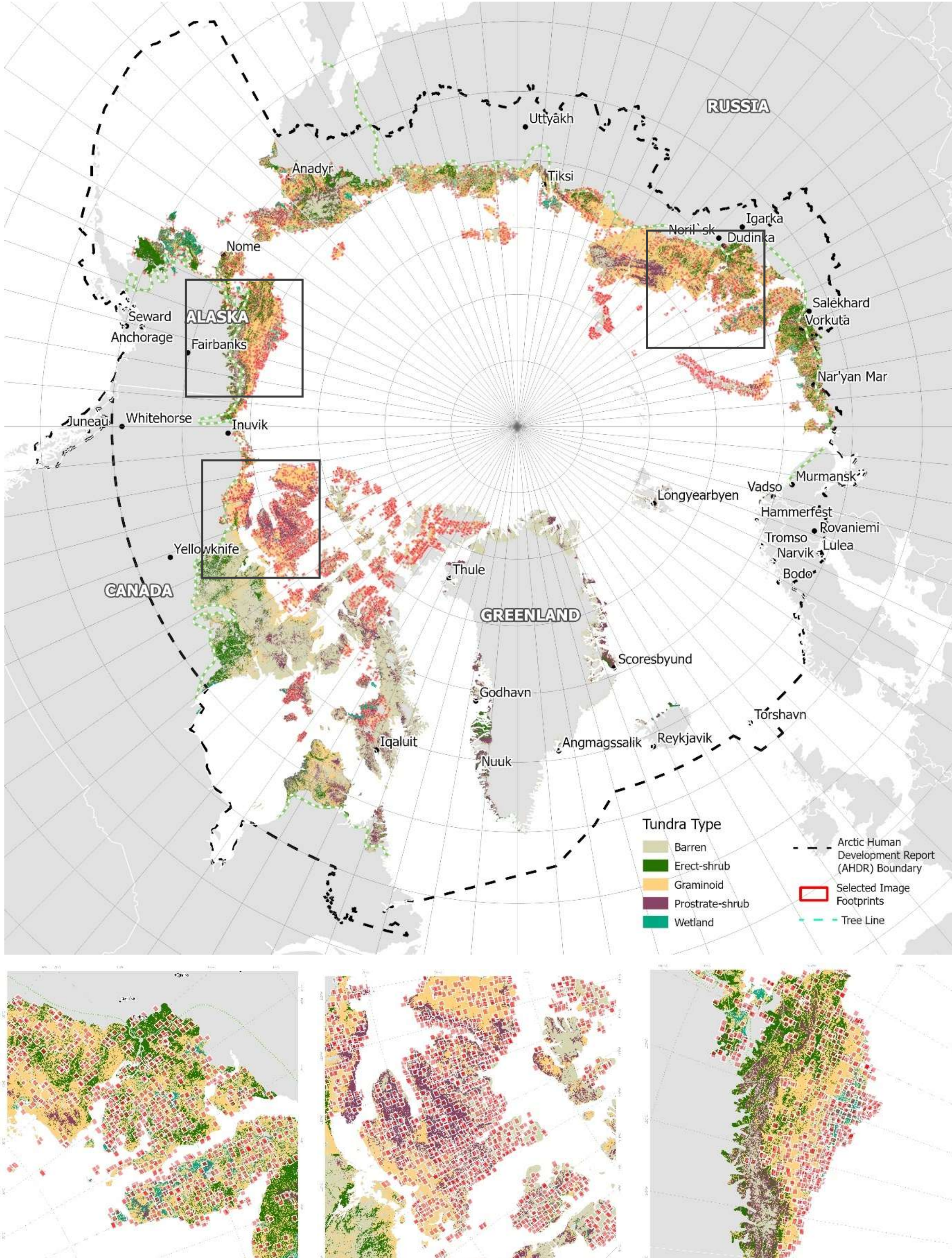

**Figure 6** Spatial distribution of selected image footprints across the Arctic tundra biome. Selected image footprints are overlaid on a Circumpolar Arctic Vegetation Map (CAVM)-derived tundra-type base layer [33], [34], including the Arctic Human Development Report (AHDR) boundary and the northern tree line to provide geographic context for the selected images.

Northslope, Alaska (160⁰ 50′ 47′W, 70⁰ 9′ 32″N)
Nunavut Territory, Canada (117⁰ 55′ W, 68⁰ 5′ 18″N)
Diksonsky District, Russia (91⁰ 37′ 8″E, 73⁰ 37′ 20″N)
Exemplar clusters
Exmeplar clusers overlain on tundra vegetaion map
Exmeplar clusers overlain on base map
Image footprint
Selected chip
Barren
Erect-shrub
Graminoid
Prostrate-shrub
Water
Wetland

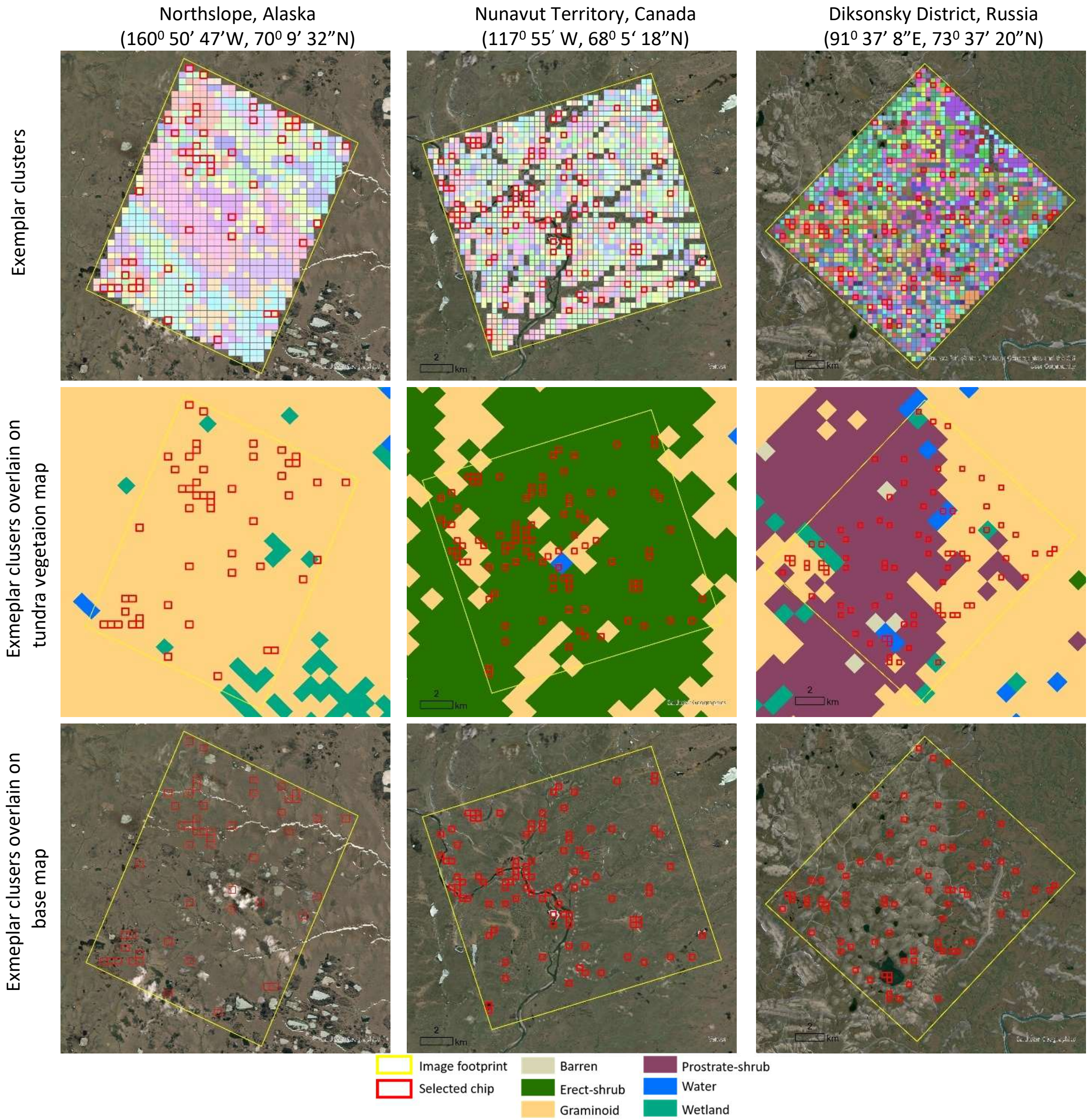

**Figure 7** Examples of chip-level clustering and exemplar-chip selection from a selected scenes with diverse tundra vegetation. Within scene grid represents candidate image chips within the scene footprint, and red outlines indicate selected exemplar chips. This second-stage clustering step reduces repeated sampling of similar local textures while preserving variation across vegetation, surface-cover, and terrain conditions.

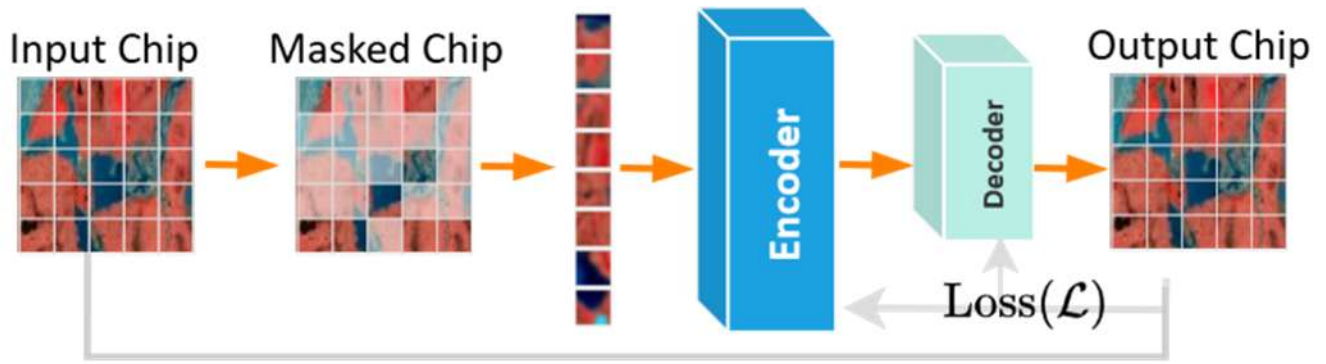

**Figure 8** Masked Auto Encoder Pretraining with domain relevant Loss function

$$\mathcal{L}_{spectral} = \mathcal{L}_{maha} + \lambda_{sam}\mathcal{L}_{SAM} \quad (1)$$

$$\mathcal{L}_{maha}(i) = (x_i - \widehat{x_i})^T \Sigma^{-1} (x_i - \widehat{x_i}) \quad (2)$$

$$\mathcal{L}_{SAM}(i) = \arccos\left(\frac{x_i^T \widehat{x_i}}{\|x_i\|_2 \|\widehat{x_i}\|_2 + \varepsilon}\right) \quad (3)$$

### D. *Downstream Task Assessments*

The MAE-pretrained ViT-Large backbone was subsequently fine-tuned on four hand-labeled Arctic feature-detection datasets to evaluate the transferability of the learned representations. For downstream adaptation, the pretrained encoder was integrated into the ViTDetLoc framework [45], which incorporates geographic location embeddings to provide spatial context during prediction. The overall detection and segmentation architecture, including the location-embedding design, follows our previously published work [45] (

Figure **9**), while the present study evaluates the effect of Arctic-specific MAE pretraining for building a foundation model for downstream performance. Each dataset was used for supervised fine-tuning and model evaluation using its corresponding task annotations, enabling assessment across multiple Arctic feature classes and scene conditions. Fine-tuning was monitored using the validation set, while the corresponding held-out test set, which was not used during training, was used for final performance reporting.

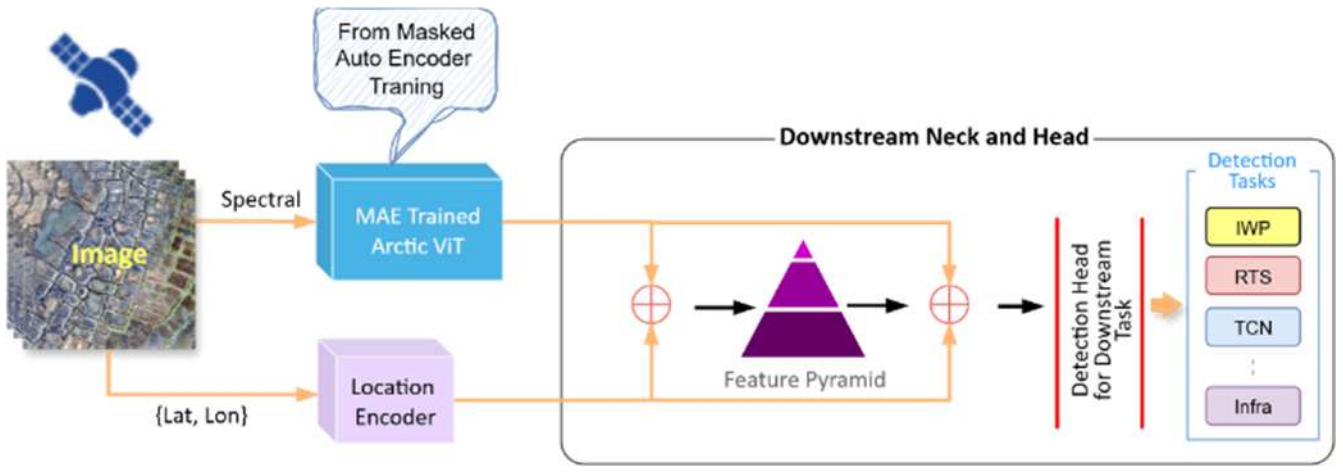

**Figure 9** Downstream task assessment pipeline used to evaluate the Arctic-pretrained ViT encoder. The pretrained ViT backbone, trained on Arctic imagery, replaces the baseline ViT backbone initialized with ImageNet-pretrained weights [45]

For comparison Prithvi-EO-2.0 [24] was selected as the primary geospatial foundation model because it represents a recent, openly available Earth observation foundation model built specifically for remote sensing transfer learning. The Prithvi model is based on a Vision Transformer architecture pretrained using a masked autoencoder objective, and its newer variants incorporate temporal and location information through spatiotemporal patch/positional embeddings and metadata-based geolocation and acquisition date embeddings. Prithvi was pretrained on 4.2 million global timeseries samples from NASA Harmonized Landsat and Sentinel-2 imagery at 30 m resolution, making it a strong general-purpose EO foundation model baseline [24]. We included Prithvi-EO-2.0 despite the resolution difference from our 0.5 m Vantor imagery because the model has been reported [24] to transfer across GEO-Bench tasks spanning different domains and spatial resolutions, including high-resolution settings from 0.1 m to 15 m. Therefore, comparison with Prithvi-EO-2.0 provides a meaningful test of whether a broad geospatial foundation model with temporal and location-aware design transfers effectively to specialized Arctic very-high-resolution feature-mapping tasks, relative to a model pretrained directly on curated Arctic Vantor imagery.

## III. Results

### E. *Clustering for sampling*

The main objective of the clustering-based curation pipeline was to extract a diverse subset of approximately 3 million training patches from 267 TB of Vantor imagery covering roughly 5 million km² of Arctic permafrost tundra. The selected subset was designed to reduce redundancy from visually repetitive and low-information areas while preserving broad scene diversity across the study domain. To evaluate this curation strategy, the exemplar-selected scenes were compared with an equally sized random subset drawn from the same archive. Distributional similarity to the full archive was measured using per-variable Wasserstein distances [57], within-subset redundancy was assessed using nearest-neighbor distances in standardized descriptor space, and descriptor-space coverage was examined using PCA [58]. These analyses allowed us to compare random and clustering-based sampling in terms of archive representativeness, redundancy reduction, and descriptor-space diversity.

Wasserstein-distance comparisons shows that the random subset more closely matched the marginal descriptor distributions of the full archive across the spectral variables, acquisition metadata, and combined 13-variable descriptor set. Wasserstein-distance comparisons showed that the random subset had lower mean distances to the full archive than the clustered subset across spectral variables, acquisition metadata, and the combined 13-variable descriptor set. These results indicate that, under the Wasserstein-distance criterion, the random subset more closely matched the marginal descriptor distributions of the full archive. This trade-off is summarized in Table 1, and discussed later, where lower Wasserstein values indicate closer agreement with the full archive distribution.

Table 1 Wasserstein-distance comparison

| **Metric group** | Mean Wasserstein | | Closer to Archive |
|---|---|---|---|
| | Random | Cluster | |
| Spectral only | 0.146 | 0.518 | Random |
| Metadata only | 0.032 | 0.119 | Random |
| All 13 variables | 0.102 | 0.365 | Random |

However, the nearest-neighbor distance analysis showed the opposite pattern. The clustered subset had larger mean, median, and interquartile nearest-neighbor distances than the random subset, indicating that the selected exemplars were more dispersed in standardized descriptor space. As shown in Figure 10, the clustered subset distribution is shifted toward larger nearest-neighbor distances, supporting the interpretation that clustering increased diversity among selected scenes.

The PCA projection (Figure 11) provided a visual summary of this representativeness–diversity trade-off. The random subset was concentrated near the densest regions of the full archive, consistent with its lower Wasserstein distances and closer agreement with the dominant archive distribution. In contrast, the clustered subset occupied a broader portion of descriptor space, including less densely populated regions, consistent with its larger nearest-neighbor distances. This pattern shows that the clustered subset provided broader descriptor-space coverage than the random subset.

### F. *Pretraining Loss*

MAE pretraining was conducted through multiple training runs with different configuration settings and checkpoint selections. The loss curve shown in **Figure 12** represents a typical training run used to assess optimization behavior. Across runs, the training loss followed a similar pattern, with a rapid decrease during the initial epochs followed by more gradual convergence. This trend indicates stable optimization of the masked reconstruction objective on the curated Arctic image-patch dataset. Selected checkpoints from these runs were then evaluated on downstream tasks, and the checkpoint with the strongest transfer performance was used for the final comparisons.

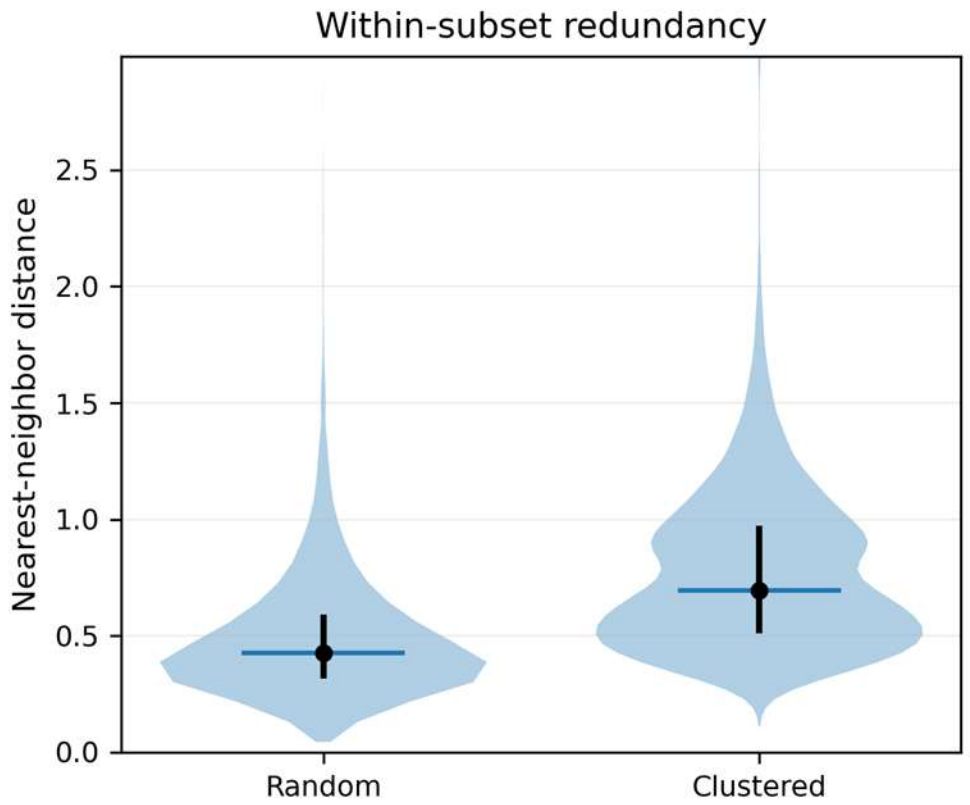

**Figure 10** Distribution of nearest-neighbor distances within the random and clustered subsets in standardized descriptor space. Higher values indicate lower within-subset redundancy. The y-axis is cropped at the 99.9th percentile for readability.

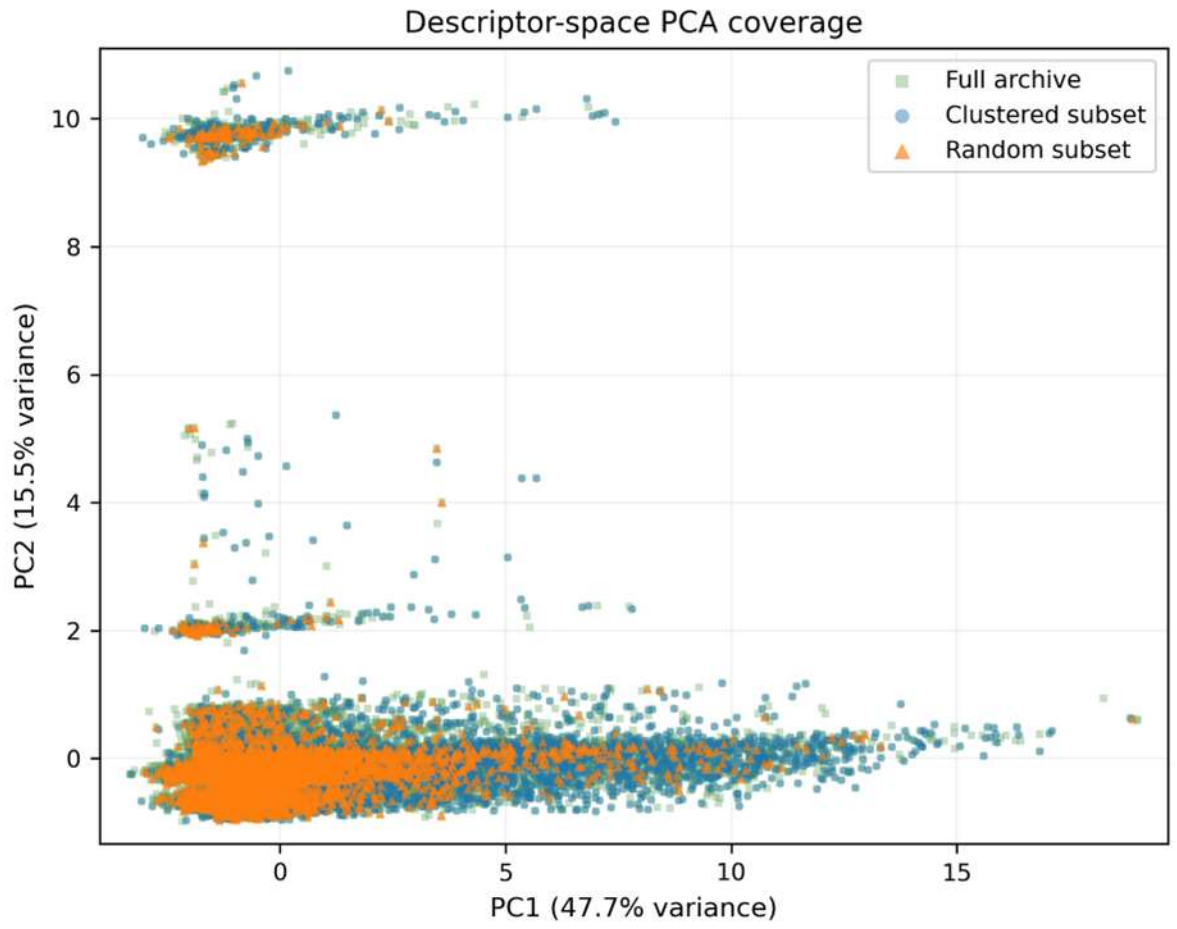

**Figure 11** PCA projection of the standardized descriptor space for the full archive (■), random subset (▲), and clustered subset (●). The random subset is concentrated near the densest region of the full archive, while the clustered subset extends across a broader portion of descriptor space, including lower-density regions. This pattern illustrates the trade-off between distributional similarity to the full archive and diversity-oriented exemplar selection.

## G. *Computing*

The two-stage clustering workflow required substantial high-performance computing resources. The first-stage affinity-propagation clustering was executed on a Frontera NVDIMM node with 2 TB of memory and 128 CPU cores. For the image collection used in this study, and using precomputed image-level statistics, this stage required approximately 40–48 hours depending on the clustering parameters. The image-scene statistics used as input to this stage were precomputed separately using a distributed data-parallel workflow across multiple nodes.

In affinity propagation, the number of exemplars was primarily controlled by the preference parameter, while the damping factor was used to stabilize convergence of the message-passing updates. As described earlier, the clustering phase was repeated several times to identify a preference value that produced the desired number of clusters and, consequently, the required number of image patches for MAE pretraining. Certain preference values resulted in too few or too many clusters relative to the target data volume, while others affected convergence stability. The final preference value was therefore selected empirically to achieve the target patch count while maintaining stable clustering behavior.

The second-stage clustering was executed in a distributed manner, with each node processing a subset of the exemplar scenes. Clustering each exemplar scene required approximately 2–11 minutes, with an average runtime of about 5 minutes per scene.

The MAE pretraining stage required multi-node GPU training and was executed on nodes with 4 x NVIDIA Quadro RTX 5000 GPUs with 16 GB of memory. Training was distributed across 16 nodes, corresponding to the maximum node count permitted for the queue, and each job was constrained by a 48-hour wall-time limit that allowed to complete approximately 60 epochs before termination. Therefore, checkpointing was used to save intermediate model states and resume training across multiple job submissions. Using this stop-and-resume workflow, MAE pretraining was continued until the model completed 800 epochs on the curated Arctic image-patch dataset. This checkpoint-based continuation was essential for completing long-running self-supervised pretraining within the queue wall-time limits.

Several MAE checkpoints from different pretraining configurations were evaluated on downstream validation tasks. The checkpoint that produced the strongest overall transfer performance was selected as the final pretrained encoder for subsequent experiments.

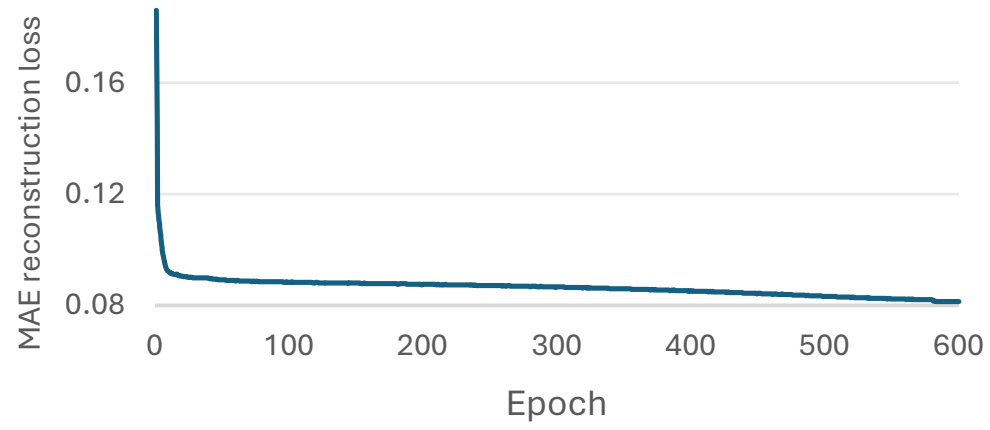

**Figure 12** Representative MAE training loss curve over epochs. The loss decreased rapidly during the initial epochs and then continued to decline gradually, indicating stable optimization of the masked reconstruction objective. Multiple MAE training runs were conducted with different configurations; this curve shows a typical run.

### *H. Downstream Analysis*

We evaluated the transferability of the Arctic MAE-pretrained encoder on the targeted downstream Arctic feature-mapping tasks. Performance was compared (Figure 13) against both the baseline model and Prithvi-EO-2.0 using foreground mean F1 score, which excludes the background class and therefore provides a more task-relevant measure of feature-detection performance. Across all evaluated tasks, the proposed MAE-pretrained model outperformed the baseline by approximately 4–5 percentage points. The MAE model also achieved higher mF1 than Prithvi-EO-2.0 in all cases, indicating that domain-specific self-supervised pretraining on curated Arctic VHSR imagery provided more transferable representations for these Arctic mapping tasks than the general-purpose remote sensing foundation model. While Prithvi-EO-2.0 provides a strong general-purpose foundation model comparison, the proposed MAE model achieved higher task-specific transfer performance. The possible reasons for this difference, including resolution mismatch and Arctic domain shift, are discussed later.

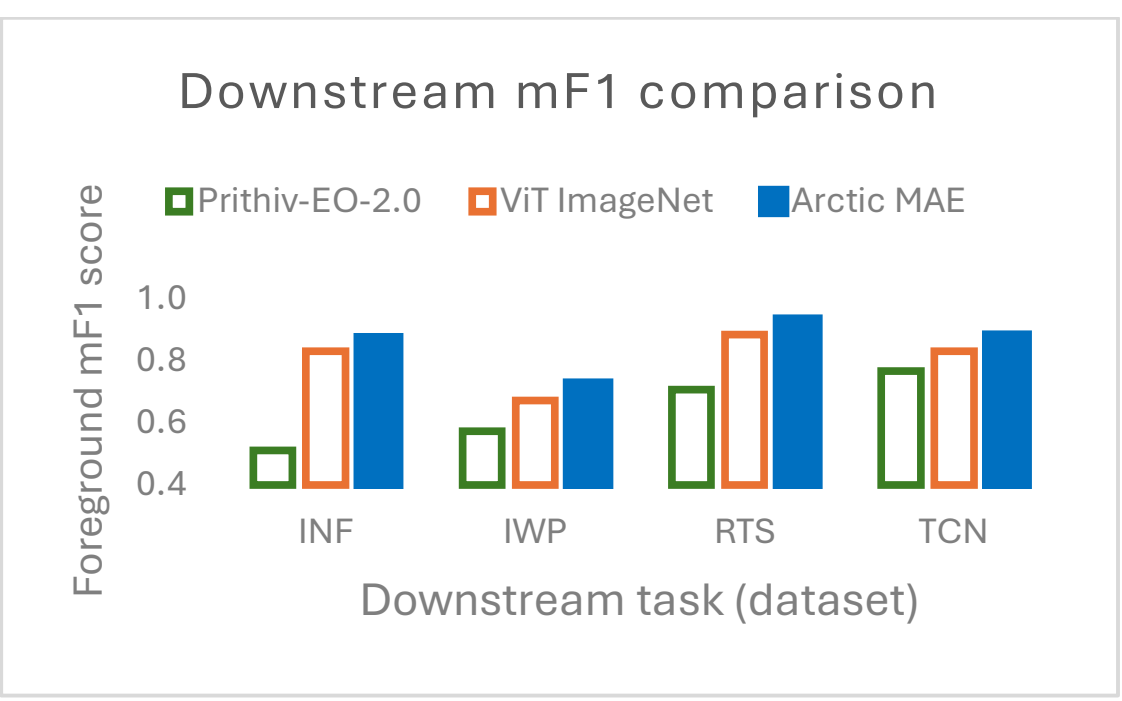

**Figure 13** Downstream task performance comparison using foreground mean F1 score. The Arctic MAE-pretrained ViT achieved the highest foreground mF1 across all evaluated datasets, outperforming both the ImageNet-initialized baseline and Prithvi-EO-2.0.

We present a compilation of illustrative examples from the respective datasets with the original image chips in false color composite and the annotated ground truth for the three datasets used in this work. We also discuss the performance of the model on the respective image chips and the possible contribution from the Arctic MAE pretraining; these qualitative examples are intended for discussion only, as quantitative model comparisons are based on the per-task metrics reported earlier.

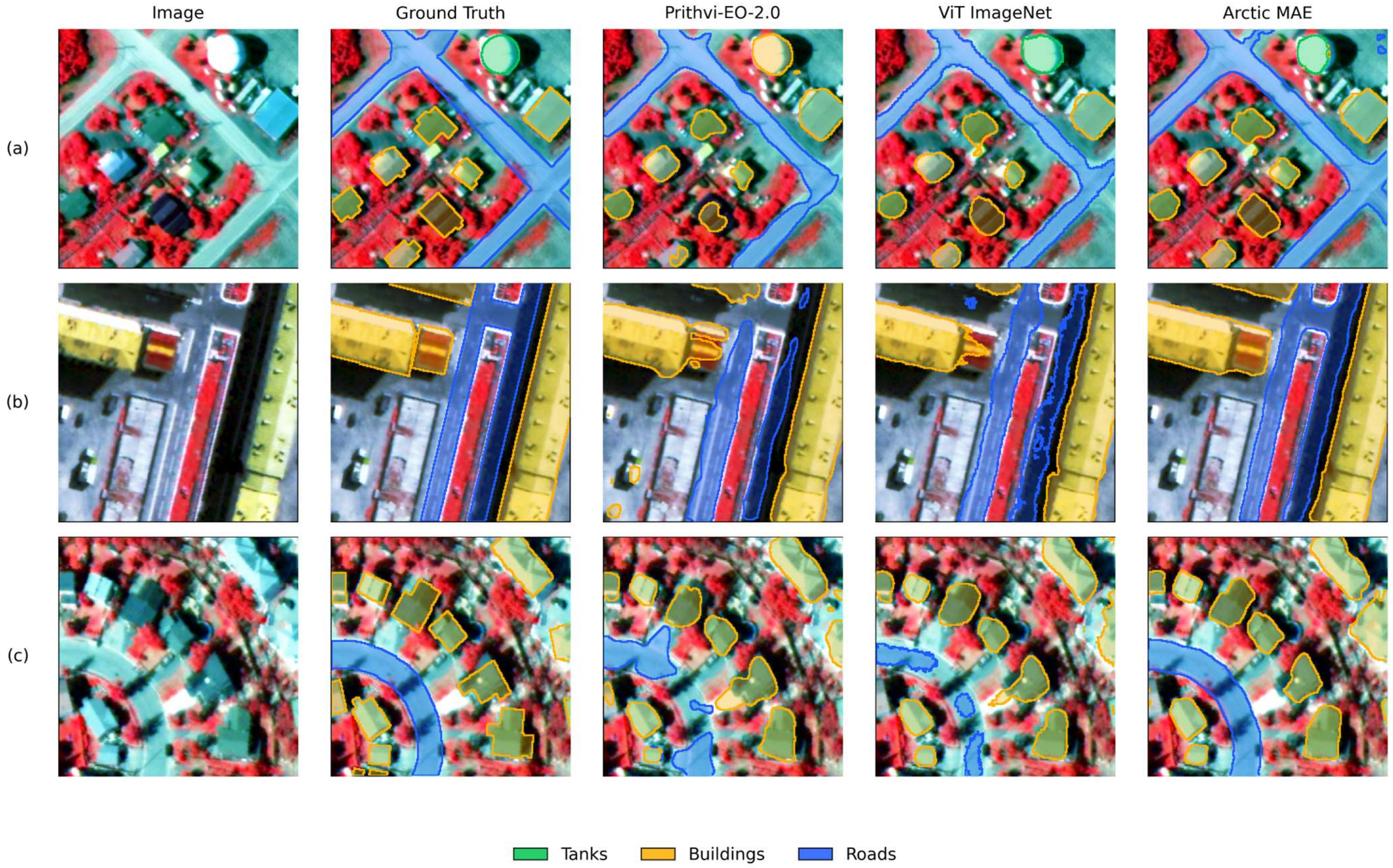

**Figure 14** Infrastructure segmentation example chips (a-c) showing base image (© Ventor), annotated ground truth, and outputs from the models. Overall, Arctic MAE, outperform the other two models on all cases with increased accuracy in segmentation matching the ground truth (a) Prithvi-EO-2.0 has misidentified the tank as a building, (b) The ViT ImageNet shows non continuous segmentation clearly indicating the inability of the encoder to capture the subtle variations on the surface due to pretraining done on natural images. (c) The Arctic MAE clearly shows a better detection compare to the other two models that have a very poor detection of the road.

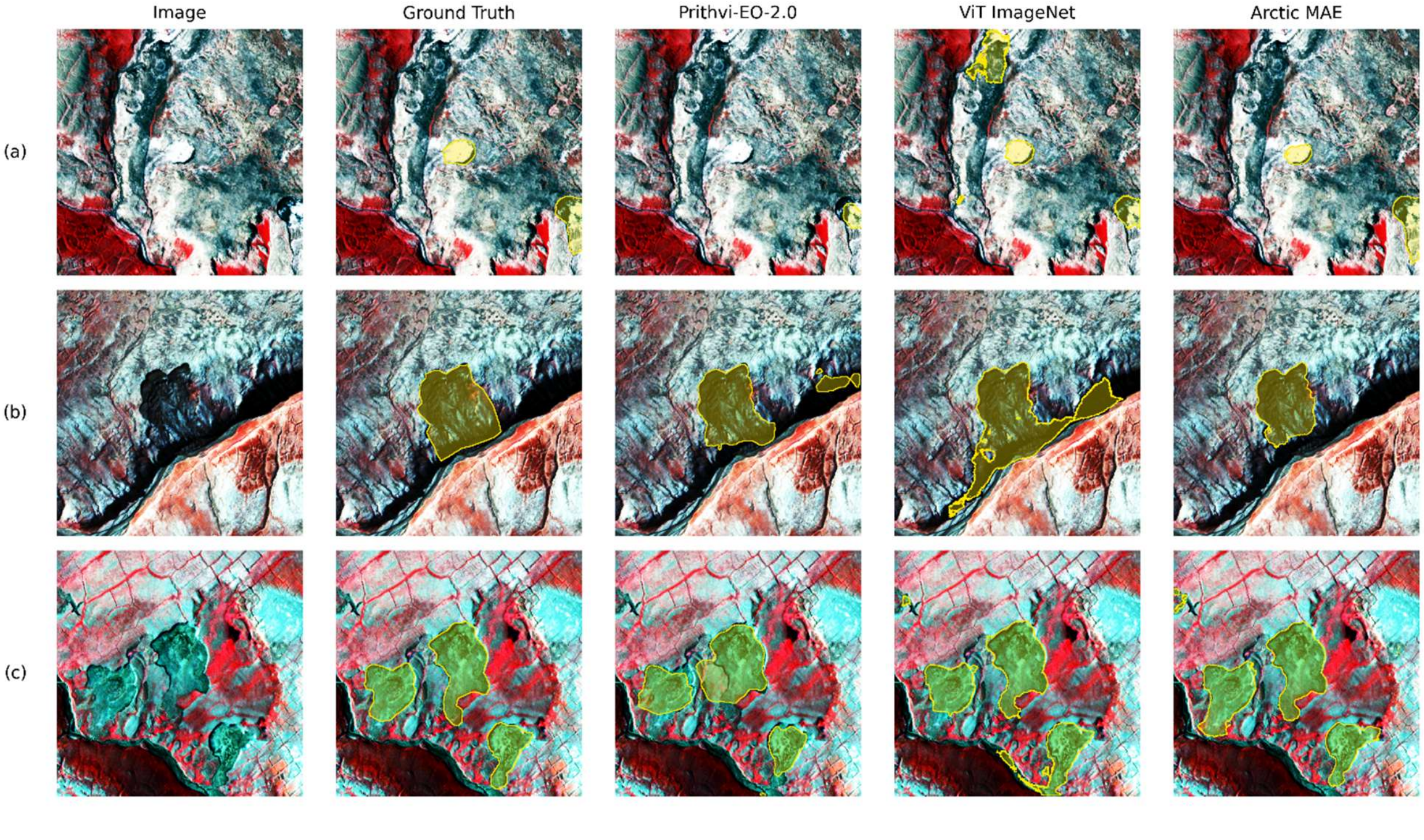

**Figure 15** RTS segmentation example chips (a-c) showing base image (© Ventor), annotated ground truth, and outputs from the models. Overall, Arctic MAE, outperforms the other two models on all cases. (a) Arctic MAE picks both objects, where Prithvi-EO-2.0 misses one object and ViT ImageNet has multiple FPs. (b) Arctic MAE detects the TP whereas the other two models picks the TP in addition to extra FPs. (c) All 3 models detect the TPs in addition to a few FPs.

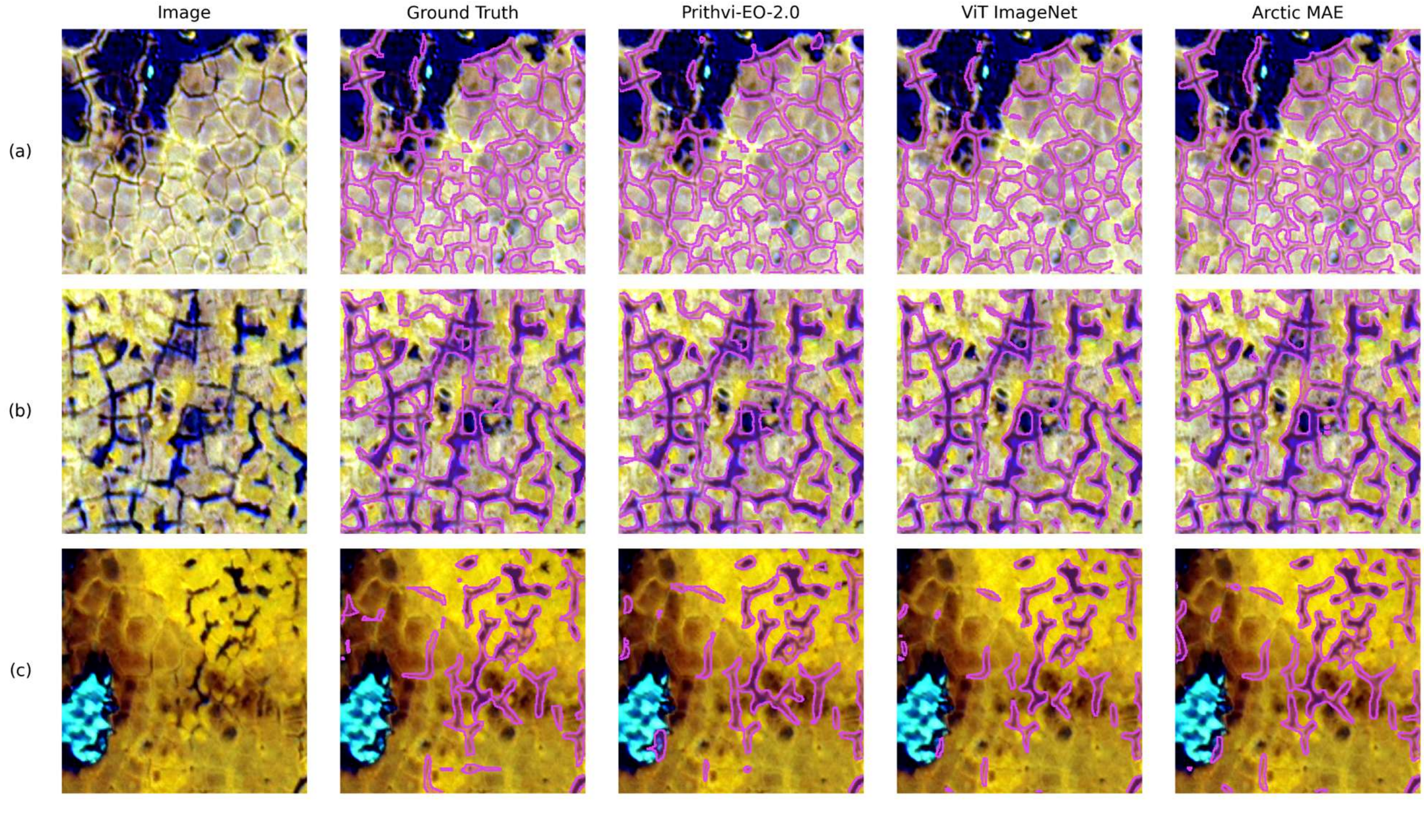

**Figure 16** TCN segmentation example chips (a-c) showing base image (© Vantor), annotated ground truth, and outputs from the models. Overall, All three models perform reasonably well although Arctic MAE is able to have more connected features compared to the other two models.

## IV. Discussion

The clustering results highlight an important trade-off in large-scale pretraining data curation. Random sampling produced subsets that more closely matched the marginal descriptor distributions of the full archive, as shown by the lower Wasserstein distances across spectral, metadata, and combined descriptor groups. However, this distributional fidelity primarily reflects the dominant conditions in the archive and does not necessarily reduce redundancy. In contrast, the clustering-based subset intentionally shifted away from the most common descriptor values, resulting in higher Wasserstein distances but greater dispersion in descriptor space. This pattern indicates that the curation pipeline prioritized diversity and redundancy reduction over simply reproducing the full archive distribution.

This interpretation is supported by the nearest-neighbor and PCA analyses, which showed that the clustered subset was more dispersed than the random subset and occupied a broader portion of descriptor space. Together, these results suggest that clustering improved coverage of diverse spectral and acquisition regimes, while sacrificing some fidelity to the dominant archive distribution. For self-supervised pretraining, this trade-off is desirable because the objective is not only to match the archive distribution, but also to expose the encoder to a broad range of visual and acquisition conditions.

This diversity-oriented clustering strategy also differs from stratified sampling approaches based on coarse land-use/land-cover and ecoregion labels, such as those used in broad geospatial foundation models. For example, Prithvi-EO-2.0 sampled global HLS training data using land-cover and ecoregion information derived from Copernicus 100 m land-cover labels and RESOLVE ecoregions, with the model pretrained on 30 m Harmonized Landsat–Sentinel imagery [24]. Such stratification is well suited for ensuring global thematic and ecological coverage, but it may not fully capture the fine-scale visual variability that is important in 0.5 m Arctic imagery. At very high spatial resolution, many relevant differences occur within the same coarse land-cover class, including vegetation texture, moisture gradients, patterned-ground structure, thaw features, shadows, and acquisition-related variability. By clustering native-resolution spectral, metadata, and chip-level texture descriptors, our curation approach selects examples based on the actual image characteristics seen by the model, rather than relying only on external coarse-resolution thematic strata. This may make the resulting pretraining corpus better aligned with fine-scale Arctic feature-mapping tasks.

Our descriptor-driven curation strategy is particularly relevant for VHSR Arctic imagery, where important variation may occur within the same coarse land-cover class. Unlike stratified sampling based on external land-cover or ecoregion labels, clustering directly uses the spectral, acquisition, and texture characteristics of the imagery seen by the model during pretraining.

The downstream results further demonstrate that this diversity-oriented curation strategy, combined with domain-specific MAE pretraining, improved the transferability of vision transformer representations for Arctic remote sensing applications.

The proposed approach consistently improved downstream foreground mF1 compared with the ImageNet-initialized baseline, which relies on representations learned from natural-image datasets rather than Earth observation imagery. The gains of approximately 5–8 F1 points are important because the evaluated tasks involve sparse, fine-scale Arctic features, including patterned ground, thaw-related landforms, tundra capillary networks, and infrastructure elements that are often spatially fragmented and visually subtle. These improvements suggest that curated Arctic MAE pretraining produced representations better suited to the visual structure of Arctic remote sensing targets, rather than simply improving background separation.

Our results further strengthen the case that remote sensing foundation models should be adapted not only to spectral and spatial characteristics, but also to the landscape dynamics of the ecoregion being modelled. Tundra Capillary Networks (TCNs, for example, may be expressed as sub-meter microtopographic trough depressions, while retrogressive thaw slumps can occur as a hillslope scale disturbance spanning several kilometers of area. Their appearance can also be strongly conditioned by seasonal and phenological surface expression. A TCN may appear as a snow-filled depression, A water-filled trough or pond structure, or even a vegetated expression dependent on factors including acquisition timing of the imagery as well as local hydrological and geological systems state. Similarly, retrogressive thaw slumps vary in their spectral and textural expression as sediment, moisture conditions, and revegetation can change overtime. These examples illustrate that Arctic remote sensing targets are not fixed visual objects; instead, they act as process and system-linked landscape features whose image signatures emerge from interactions between spatial scale, microtopographic patterning, surface moisture, vegetation, geology, and acquisition window of the imagery. Ecoregion-specific pretraining therefore emerges as a valuable technique because it exposes the encoder to the range of characteristics within the various Arctic surface expressions that govern downstream appearance of features. The consistent gains in performance across sub-meter TCNs and meter-kilometer scale RTS features suggest that Arctic-focused RSFMs can learn representations more closely aligned with tundra landscape processes than models that are pre-trained on globally aggregated or coarser-image distributions.

The comparison with Prithvi-EO-2.0 further highlights the importance of domain-specific pretraining. Although Prithvi-EO-2.0 is a strong general-purpose Earth observation foundation model and has reported transfer across multiple benchmark settings, its pretraining data differs substantially from our downstream setting in both spatial resolution and image domain. Prithvi-EO-2.0 was pretrained primarily on medium-resolution satellite imagery, whereas our MAE model was pretrained directly on 0.5 m Arctic Vantor imagery. The higher performance of the proposed model suggests that cross-resolution transfer from a general-purpose foundation model may not be sufficient for specialized Arctic feature-mapping tasks where local texture, fine spatial detail, and sparse class distributions are critical.

## V. Conclusion

This study demonstrates the value of domain-specific self-supervised pretraining for Arctic remote sensing applications. By combining diversity-aware image curation with MAE pretraining on 0.5 m Ventor imagery, we developed an Arctic-focused vision transformer encoder that improved downstream feature-mapping performance across multiple tasks. The observed gains in foreground mF1 indicate that the proposed pretraining strategy helped the model learn representations that are better suited for sparse, fine-scale, and visually subtle Arctic surface features.

The results also highlight the importance of matching foundation model pretraining data to the target domain. Although general-purpose remote sensing foundation models provide a valuable starting point, our comparison with Prithvi-EO-2.0 suggests that cross-resolution and cross-domain transfer may be limited when downstream tasks depend on very-high-resolution texture, local morphology, and Arctic-specific surface conditions. In this context, curated domain-specific pretraining provided more effective transferable representations than relying on a broader foundation model alone.

More broadly, this work shows that large archives of unlabeled Arctic imagery can be used to build useful foundation representations for environmental mapping. The proposed workflow provides a scalable path for transforming very-high-resolution satellite image collections into pretrained models that support downstream detection and segmentation tasks with limited labeled data.

This study demonstrates the value of domain-specific self-supervised pretraining for Arctic remote sensing applications. By combining diversity-aware image curation with MAE pretraining on 0.5 m Vantor imagery, we developed an Arctic-focused vision transformer encoder that improved downstream feature-mapping performance across multiple tasks. The observed gains in foreground mF1 of approximately 5–8 percentage for all tested downstream tasks indicate that the proposed pretraining strategy helped the model learn representations better suited to sparse, fine-scale, and visually subtle Arctic surface features.

The results also highlight the importance of aligning foundation-model pretraining data with the target domain. Although general-purpose remote sensing foundation models provide a valuable starting point, our comparison with Prithvi-EO-2.0 suggests that cross-resolution and cross-domain transfer may be limited when downstream tasks depend on very-high-resolution texture, local morphology, and Arctic-specific surface conditions. In this setting, curated domain-specific pretraining provided more effective transferable representations than relying on a broader foundation model alone.

More broadly, this work shows that large archives of unlabeled Arctic imagery can be transformed into useful pretrained representations for environmental mapping. The proposed workflow provides a scalable path for converting very-high-resolution satellite image collections into domain-specialized foundation models that support downstream detection and segmentation tasks with limited labeled data.

## References

[1] J. D. Phillips, "Evolutionary geomorphology: thresholds and nonlinearity in landform response to environmental change," *Hydrol. Earth Syst. Sci.*, vol. 10, no. 5, pp. 731–742, Oct. 2006, doi: 10.5194/hess-10-731-2006.

[2] N. Wunderling, J. F. Donges, J. Kurths, and R. Winkelmann, "Interacting tipping elements increase risk of climate domino effects under global warming," *Earth Syst. Dyn.*, vol. 12, no. 2, pp. 601–619, Jun. 2021, doi: 10.5194/esd-12-601-2021.

[3] C. L. E. Franzke *et al.*, "Perspectives on tipping points in integrated models of the natural and human Earth system: cascading effects and telecoupling," *Environ. Res. Lett.*, vol. 17, no. 1, p. 015004, Jan. 2022, doi: 10.1088/1748-9326/ac42fd.

[4] C. Vaduva, M. Iapaolo, and M. Datcu, "A Scientific Perspective on Big Data in Earth Observation," in *Principles of Data Science*, H. R. Arabnia, K. Daimi, R. Stahlbock, C. Soviany, L. Heilig, and K. Brüssau, Eds., Cham: Springer International Publishing, 2020, pp. 155–188. doi: 10.1007/978-3-030-43981-1_8.

[5] A. C. Foster *et al.*, "Disturbances in North American boreal forest and Arctic tundra: impacts, interactions, and responses," *Environ. Res. Lett.*, vol. 17, no. 11, p. 113001, Oct. 2022, doi: 10.1088/1748-9326/ac98d7.

[6] L. Filchev, L. Pashova, V. Kolev, and S. Frye, "Challenges and Solutions for Utilizing Earth Observations in the 'Big Data' era," Dec. 2018, doi: 10.5281/zenodo.2391937.

[7] M. A. Wulder *et al.*, "Current status of Landsat program, science, and applications," *Remote Sens. Environ.*, vol. 225, pp. 127–147, May 2019, doi: 10.1016/j.rse.2019.02.015.

[8] N. Sisodiya, N. Dube, O. Prakash, and P. Thakkar, "Scalable big earth observation data mining algorithms: a review," *Earth Sci. Inform.*, vol. 16, no. 3, pp. 1993–2016, Sep. 2023, doi: 10.1007/s12145-023-01032-5.

[9] G. Gao, L. Yao, W. Li, L. Zhang, and M. Zhang, "Onboard Information Fusion for Multisatellite Collaborative Observation: Summary, challenges, and perspectives," *IEEE Geosci. Remote Sens. Mag.*, vol. 11, no. 2, pp. 40–59, Jun. 2023, doi: 10.1109/MGRS.2023.3274301.

[10] E. Dritsas and M. Trigka, "Remote Sensing and Geospatial Analysis in the Big Data Era: A Survey," *Remote Sens.*, vol. 17, no. 3, p. 550, Jan. 2025, doi: 10.3390/rs17030550.

[11] D. Tuia *et al.*, "Artificial Intelligence to Advance Earth Observation: A review of models, recent trends, and pathways forward," *IEEE Geosci. Remote Sens. Mag.*, vol. 13, no. 4, pp. 119–141, Dec. 2025, doi: 10.1109/MGRS.2024.3425961.

[12] K. He, X. Chen, S. Xie, Y. Li, P. Dollar, and R. Girshick, "Masked Autoencoders Are Scalable Vision Learners," in *2022 IEEE/CVF Conference on Computer Vision and Pattern Recognition (CVPR)*, New Orleans, LA, USA: IEEE, Jun. 2022, pp. 15979–15988. doi: 10.1109/CVPR52688.2022.01553.

[13] Z. Zhou and X. Liu, "Masked Autoencoders in Computer Vision: A Comprehensive Survey," *IEEE Access*, vol. 11, pp. 113560–113579, 2023, doi: 10.1109/ACCESS.2023.3323383.

[14] A. Dosovitskiy *et al.*, "An Image is Worth 16x16 Words: Transformers for Image Recognition at Scale," in *9th International Conference on Learning Representations, ICLR 2021, Virtual Event, Austria, May 3-7, 2021*, OpenReview.net, 2021. Accessed: Nov. 15, 2024. [Online]. Available: https://openreview.net/forum?id=YicbFdNTTy

[15] C. Burnett and T. Blaschke, "A multi-scale segmentation/object relationship modelling methodology for landscape analysis," *Ecol. Model.*, vol. 168, no. 3, pp. 233–249, Oct. 2003, doi: 10.1016/S0304-3800(03)00139-X.

[16] L. Drăguţ and C. Eisank, "Object representations at multiple scales from digital elevation models," *Geomorphology*, vol. 129, no. 3, pp. 183–189, Jun. 2011, doi: 10.1016/j.geomorph.2011.03.003.

[17] G. J. Hay, "Visualizing Scale-Domain Manifolds: A Multiscale Geo-Object-Based Approach," in *Scale Issues in Remote Sensing*, John Wiley & Sons, Ltd, 2014, pp. 139–169. doi: 10.1002/9781118801628.ch08.

[18] T. Blaschke, "Object based image analysis for remote sensing," *ISPRS J. Photogramm. Remote Sens.*, vol. 65, no. 1, pp. 2–16, Jan. 2010, doi: 10.1016/j.isprsjprs.2009.06.004.

[19] G. Castilla and G. J. Hay, "Image objects and geographic objects," in *Object-Based Image Analysis: Spatial Concepts for Knowledge-Driven Remote Sensing Applications*, T. Blaschke, S. Lang, and G. J. Hay, Eds., Berlin, Heidelberg: Springer, 2008, pp. 91–110. doi: 10.1007/978-3-540-77058-9_5.

[20] T. Blaschke *et al.*, "Geographic Object-Based Image Analysis - Towards a new paradigm," *ISPRS J. Photogramm. Remote Sens. Off. Publ. Int. Soc. Photogramm. Remote Sens.*, vol. 87, no. 100, pp. 180–191, Jan. 2014, doi: 10.1016/j.isprsjprs.2013.09.014.

[21] D. Tuia, C. Persello, and L. Bruzzone, "Domain Adaptation for the Classification of Remote Sensing Data: An Overview of Recent Advances," *IEEE Geosci. Remote Sens. Mag.*, vol. 4, no. 2, pp. 41–57, Jun. 2016, doi: 10.1109/MGRS.2016.2548504.

[22] X. X. Zhu *et al.*, "Deep Learning in Remote Sensing: A Comprehensive Review and List of Resources," *IEEE Geosci. Remote Sens. Mag.*, vol. 5, no. 4, pp. 8–36, Dec. 2017, doi: 10.1109/MGRS.2017.2762307.

[23] P. Ghamisi *et al.*, "Multisource and Multitemporal Data Fusion in Remote Sensing: A Comprehensive Review of the State of the Art," *IEEE Geosci. Remote Sens. Mag.*, vol. 7, no. 1, pp. 6–39, Mar. 2019, doi: 10.1109/MGRS.2018.2890023.

[24] D. Szwarcman *et al.*, "Prithvi-EO-2.0: A Versatile Multitemporal Foundation Model for Earth Observation Applications," *IEEE Trans. Geosci. Remote Sens.*, vol. 64, pp. 1–20, 2026, doi: 10.1109/TGRS.2025.3642610.

[25] J. Jakubik *et al.*, "Foundation Models for Generalist Geospatial Artificial Intelligence," Nov. 08, 2023, *arXiv*: arXiv:2310.18660. doi: 10.48550/arXiv.2310.18660.

[26] Y. Cong *et al.*, "SatMAE: pre-training transformers for temporal and multi-spectral satellite imagery," in *Proceedings of the 36th International Conference on Neural Information Processing Systems*, in NIPS '22. Red Hook, NY, USA: Curran Associates Inc., Nov. 2022, pp. 197–211.

[27] C. J. Reed *et al.*, "Scale-MAE: A Scale-Aware Masked Autoencoder for Multiscale Geospatial Representation Learning," in *2023 IEEE/CVF International Conference on Computer Vision (ICCV)*, Paris, France: IEEE, Oct. 2023, pp. 4065–4076. doi: 10.1109/ICCV51070.2023.00378.

[28] "Full article: GeoAI: spatially explicit artificial intelligence techniques for geographic knowledge discovery and beyond." Accessed: May 15, 2026. [Online]. Available: https://www.tandfonline.com/doi/full/10.1080/13658816.2019.1684500

[29] E. Rolf, K. Klemmer, C. Robinson, and H. Kerner, "Position: mission critical - satellite data is a distinct modality in machine learning," in *Proceedings of the 41st International Conference on Machine Learning*, in ICML'24, vol. 235. Vienna, Austria: JMLR.org, Jul. 2024, pp. 42691–42706.

[30] C. Huo *et al.*, "When Remote Sensing Meets Foundation Model: A Survey and Beyond," *Remote Sens.*, vol. 17, no. 2, p. 179, Jan. 2025, doi: 10.3390/rs17020179.

[31] Z. Huang *et al.*, "A Survey on Remote Sensing Foundation Models: From Vision to Multimodality," Mar. 28, 2025, *arXiv*: arXiv:2503.22081. doi: 10.48550/arXiv.2503.22081.

[32] A. Xiao *et al.*, "Foundation Models for Remote Sensing and Earth Observation: A survey," *IEEE Geosci. Remote Sens. Mag.*, vol. 13, no. 4, pp. 297–324, Dec. 2025, doi: 10.1109/MGRS.2025.3576766.

[33] M. K. Raynolds *et al.*, "A raster version of the Circumpolar Arctic Vegetation Map (CAVM)," *Remote Sens. Environ.*, vol. 232, p. 111297, 2019.

[34] D. A. Walker *et al.*, "The Circumpolar Arctic vegetation map," *J. Veg. Sci.*, vol. 16, no. 3, pp. 267–282, 2005, doi: 10.1111/j.1654-1103.2005.tb02365.x.

[35] A. K. Liljedahl *et al.*, "Pan-Arctic ice-wedge degradation in warming permafrost and its influence on tundra hydrology," *Nat. Geosci.*, vol. 9, p. 312, 2016.

[36] C. Witharana *et al.*, "Automated Detection of Retrogressive Thaw Slumps in the High Arctic Using High-Resolution Satellite Imagery," *Remote Sens.*, vol. 14, no. 17, Art. no. 17, Jan. 2022, doi: 10.3390/rs14174132.

[37] A. Beamish *et al.*, "Recent trends and remaining challenges for optical remote sensing of Arctic tundra vegetation: A review and outlook," *Remote Sens. Environ.*, vol. 246, p. 111872, Sep. 2020, doi: 10.1016/j.rse.2020.111872.

[38] X.-Y. Jin *et al.*, "Impacts of climate-induced permafrost degradation on vegetation: A review," *Adv. Clim. Change Res.*, vol. 12, no. 1, pp. 29–47, Feb. 2021, doi: 10.1016/j.accre.2020.07.002.

[39] M. M. P. D. Heijmans *et al.*, "Tundra vegetation change and impacts on permafrost," *Nat. Rev. Earth Environ.*, vol. 3, no. 1, pp. 68–84, Jan. 2022, doi: 10.1038/s43017-021-00233-0.

[40] J. E. Vonk *et al.*, "Reviews and syntheses: Effects of permafrost thaw on Arctic aquatic ecosystems," *Biogeosciences*, vol. 12, no. 23, pp. 7129–7167, Dec. 2015, doi: 10.5194/bg-12-7129-2015.

[41] K. R. Miner *et al.*, "Emergent biogeochemical risks from Arctic permafrost degradation," *Nat. Clim. Change*, vol. 11, no. 10, pp. 809–819, Oct. 2021, doi: 10.1038/s41558-021-01162-y.

[42] D. A. Streletskiy, S. Clemens, J.-P. Lanckman, and N. I. Shiklomanov, "The costs of Arctic infrastructure damages due to permafrost degradation," *Environ. Res. Lett.*, vol. 18, no. 1, p. 015006, Jan. 2023, doi: 10.1088/1748-9326/acab18.

[43] E. Manos, C. Witharana, and A. K. Liljedahl, "Permafrost thaw-related infrastructure damage costs in Alaska are projected to double under medium and high emission scenarios," *Commun. Earth Environ.*, vol. 6, no. 1, pp. 1–11, Mar. 2025, doi: 10.1038/s43247-025-02191-7.

[44] B. Swan, M. Laverdiere, H. L. Yang, and A. Rose, "Iterative self-organizing SCEne-LEvel sampling (ISOSCELES) for large-scale

building extraction," *GIScience Remote Sens.*, vol. 59, no. 1, pp. 1–16, Dec. 2022, doi: 10.1080/15481603.2021.2006433.

[45] A. S. Perera *et al.*, "Pan-Arctic Permafrost Landform and Human-Built Infrastructure Feature Detection With Vision Transformers and Location Embeddings," *IEEE J. Sel. Top. Appl. Earth Obs. Remote Sens.*, vol. 19, pp. 7858–7879, 2026, doi: 10.1109/JSTARS.2025.3648673.

[46] J. L. Brown, O. J. F. Jr, J. A. Heginbottom, and E. S. Melnikov, "Circum-Arctic map of permafrost and ground-ice conditions," U.S. Geological Survey, 45, 1997. doi: 10.3133/cp45.

[47] E. Manos, C. Witharana, M. R. Udawalpola, A. Hasan, and A. K. Liljedahl, "Convolutional Neural Networks for Automated Built Infrastructure Detection in the Arctic Using Sub-Meter Spatial Resolution Satellite Imagery," *Remote Sens.*, vol. 14, no. 11, Art. no. 11, Jan. 2022, doi: 10.3390/rs14112719.

[48] A. S. Perera, C. Witharana, E. Manos, and A. K. Liljedahl, "Hyperparameter Optimization for Large-Scale Remote Sensing Image Analysis Tasks: A Case Study Based on Permafrost Landform Detection Using Deep Learning," *IEEE Access*, vol. 12, pp. 43062–43077, 2024, doi: 10.1109/ACCESS.2024.3379142.

[49] M. Pimenta, C. Witharana, A. Perera, A. Liljedahl, and E. Manos, "Pan-Alaskan permafrost tundra capillary network detection and graph theoretic analysis from <1 meter resolution satellite imagery (2011 - 2025)." 2026. doi: 10.18739/A24Q7QS3R.

[50] E. Dinerstein *et al.*, "An Ecoregion-Based Approach to Protecting Half the Terrestrial Realm," *BioScience*, vol. 67, no. 6, pp. 534–545, 2017.

[51] F. A. Kruse *et al.*, "The Spectral Image Processing System (SIPS)—Interactive visualization and analysis of imaging spectrometer data," *Remote Sens. Environ.*, vol. 44, no. 2–3, pp. 145-163, 1993.

[52] H. Feilhauer, G. P. Asner, R. E. Martin, and S. Schmidtlein, "Brightness-normalized partial least squares regression for hyperspectral data," *J. Quant. Spectrosc. Radiat. Transf.*, vol. 111, no. 12–13, pp. 1947-1957, 2010, doi: 10.1016/j.jqsrt.2010.03.007.

[53] M. Fauvel, A. Villa, J. Chanussot, and J. A. Benediktsson, "Mahalanobis kernel for the classification of hyperspectral images," in *Proc. IEEE Int. Geosci. Remote Sens. Symp. (IGARSS*, Honolulu, HI, USA, 2010, pp. 3724-3727,. doi: 10.1109/IGARSS.2010.5651956.

[54] J. Krishnaswamy, K. S. Bawa, K. N. Ganeshaiah, and M. C. Kiran, "Quantifying and mapping biodiversity and ecosystem services: Utility of a multi-season NDVI based Mahalanobis distance surrogate," *Remote Sens. Environ.*, vol. 113, no. 4, pp. 857-867, 2009.

[55] W. Zhang, H. Guo, S. Liu, and S. Wu, "Attention-aware spectral difference representation for hyperspectral anomaly detection," *Remote Sens.*, vol. 15, no. 10, Art. no. 2652, 2023, doi: 10.3390/rs15102652.

[56] D. Stanzione, J. West, R. T. Evans, T. Minyard, O. Ghattas, and D. K. Panda, "Frontera: The Evolution of Leadership Computing at the National Science Foundation," in *Practice and Experience in Advanced Research Computing*, in PEARC '20. New York, NY, USA: Association for Computing Machinery, Jul. 2020, pp. 106–111. doi: 10.1145/3311790.3396656.

[57] A. Ramdas, N. Garcia, and M. Cuturi, "On Wasserstein two-sample testing and related families of nonparametric tests," *Entropy*, vol. 19, no. 2, p. 47, 2017, doi: 10.3390/e19020047.

[58] I. T. Jolliffe, *Principal Component Analysis*, 2nd ed. New York, NY, USA: Springer, 2002.